# Identification of Bivariate Causal Directionality Based on Anticipated Asymmetric Geometries

**Alex Glushkovsky**

**Abstract**

Identification of causal directionality in bivariate numerical data is a fundamental research problem with important practical implications. This paper presents two alternative methods to identify direction of causation by considering conditional distributions: (1) Anticipated Asymmetric Geometries (AAG) and (2) Monotonicity Index (MI). The AAG method compares the actual conditional distributions to anticipated ones along two variables. Different comparison metrics, such as Pearson correlation, cosine distance, Hellinger distance, Jaccard index, Jeffreys divergence, K-L divergence, K-S distance, MAE, MSE, mutual information, and Wasserstein distance have been evaluated. Anticipated distributions have been projected as normal based on dual response statistics: mean and standard deviation. The MI method compares the calculated monotonicity indexes of the gradients of conditional distributions along two axes and exhibits count of gradient sign changes. Both methods assume stochastic properties of the bivariate data and exploit anticipated unimodality of conditional distributions of the effect. The proposed methods are straightforward and include only a limited number of hyperparameters that affect the accuracy of the identification. For a given set of hyperparameters, both the AAG and MI methods provide a unique, deterministic solution. To address sensitivity to hyperparameters, tuning has been done by utilizing a full factorial Design of Experiment. It turns out that the AAG method outperforms MI, achieving top weighted accuracies of 82.7% with simple tuning and 84.4% with size-adaptive tuning, compared with 81.6% for GRCI or 82.0% for CAREFL-H on the 99 pairs of the Tubingen real-world cause-effect examples. To evaluate the decisiveness of the identification method, a decision tree was fitted on the input data's symmetrical bivariate statistics to detect misclassified cases.

## 1. Introduction

Identification of causal directionality observing only bivariate numerical data is a fundamental research problem with important practical implications. Having two numeric columns in a dataset, X and Y, the question of casual relationships arises: Does X cause Y ($X \rightarrow Y$) or, vice versa, does Y cause X ($Y \rightarrow X$)? This article is limited to cases with simple bivariate causation and does not consider other cases of dependencies, such as confounding, feedback, or selection bias.

An overview and description of the two leading methods of distinguishing cause from effect using observational data are provided in (Mooij *et al*, 2016): Additive Noise Models (ANM) and Information Geometric Causal Inference (IGCI). It was shown that the most accurate method is ANM achieving a weighted accuracy of 63 ± 10 % and an AUC of 0.74 ± 0.05 on the Tübingen real-world data (Hoyer *et al*, 2009).

ANM have been enhanced by addressing heteroscedasticity (Immer *et al*, 2023; Khemakhem *et al*, 2023; Strobl and Lasko, 2023; Xiangyu Sun and Schulte, 2023). Thus, Generalized Root Causal Inference (GRCI) and location-scale noise models with independence testing (CAREFL-H) reach 81.6% and 82.0% accuracies, correspondingly.

This paper presents alternative methods of identification of causal directionality based on Anticipated Asymmetric Geometries (AAG) and Monotonicity Index that have been evaluated on the Tübingen real-word data used in (Mooij *et al*, 2014) paper.

While not intended as a comprehensive review of the causal directionality, this paper concentrates on methods based on the 2D joint and conditional distributions of bivariate data.

## 2. Rationales of Causal Directionality Identification

Let us consider a deterministic function definition $Y=f(X)$ where for any given value of $X$, there is only a single unique value of $Y$ while the uniqueness of $X$ given $Y$ is not constrained. The latter is a case where $f(X)$ is a non-injective function and, therefore, there is no inverse function. That condition breaks the symmetry and supports causal $X \rightarrow Y$ discovery.

The principle described for deterministic functions has been extended to stochastic causal relationships that generalize most practical cases where some noises are present, such as measurement noise.

Thus, Additive Noise Models (ANM) are built based on the following key principle: if $Y=f(X)+\varepsilon$, where $\varepsilon$ is additive independent noise, while inverse function $X=g(Y)+\varepsilon'$ has noise $\varepsilon'$ dependent on Y, then casual relationship $X \rightarrow Y$ can be assumed (Hoyer *et al*, 2009; Mooij and Janzing, 2010).

Enhanced methods that address heteroscedasticity are based on dual responses $Y=f(X)+s(X)\cdot\varepsilon$, where the second function isolates magnitude of the noise $\varepsilon$ (Immer *et al*, 2023; Khemakhem *et al*, 2023; Strobl and Lasko, 2023; Xiangyu Sun and Schulte, 2023).

Dealing with stochastic bivariate cases, it is reasonable to consider probability density distributions. The fundamental work exposing joint, conditional, and marginal distributions in the context of causal relationships is provided by (Pearl, 2000). This paper focuses on the identification of causal directionality by considering conditional distributions and features the following two methods:

i. **Anticipated Asymmetric Geometries (AAG)**: This method is based on the comparison between the actual conditional distributions to anticipated ones along two axes. The anticipated distributions have been projected as normal based on estimated dual response statistics, i.e., mean and standard deviation.
ii. **Monotonicity Index**: The alternative approach that compares the calculated monotonicity indexes is based on the gradients of conditional distributions along two axes, where indexes exhibit counts of gradient sign changes.

Both methods assume stochastic properties of the bivariate data and exploit the anticipated unimodality of conditional distributions of the effect. Thus, having bivariate numerical data $X$ and $Y$, where, as an example $X$ causes $Y$, i.e., there is $X \rightarrow Y$ cause-effect relationship, then it is anticipated that the conditional distribution $\varphi_{Y|X}(Y|X)$ of the effect $Y$ follows unimodal distribution. This simple characterization indicates existence of a single factor that causes $Y$ given a value of $X$. Observing multimodal distribution $\varphi_{Y|X}(Y|X)$, for example, bimodal, assumes existence of at least one more contributing factor other than $X$. On the contrary, the conditional distribution $\varphi_{X|Y}(X|Y)$ in case of $X$ to be a cause, is not necessarily limited to an unimodal case.

## 3. Anticipated Asymmetric Geometries (AAG) Methods

Let us consider empirical 2D joint probability density distribution $\varphi(X_i, Y_j)$ that has been fitted based on original bivariate numerical data ($X$, $Y$) by applying $k \cdot k$ grid, $i, j \in [1, k]$. Practically, the *stats.gaussian_kde* from *scipy* has been used to fit 2D joint probability density distributions.

Having that 2D distribution fitted, it is possible to derive empirical conditional distributions $\varphi_{X|Y}(X_i|Y_j)$ and $\varphi_{Y|X}(Y_j\,|X_i)$. Observation of deviations of the empirical conditional distributions from anticipated unimodal geometries of the effect breaks the symmetry and supports discovery of causal directionality. Deviations can be interpreted as a loss of information.

The normally distributed noise is the simplest form of anticipated geometries ensuring (a) unimodality of the conditional distribution, (b) symmetrical distribution of noise around mean, (c) maximum entropy, i.e., noise with maximum disorder, and (d) according to The Central Limit Theorem, a convergence of many small, independent, random sources of noise. Notably, the independence property of noise aligns to the ANM approach (Hoyer *et al*, 2009).

Suppose that $X \rightarrow Y$, then observing $\varphi_{Y|X}(Y_j|X_i)$ for a given $X_i$ value, the question is: What information will be lost if the conditional distribution is encoded into a dual response, i.e., mean $\mu_Y(\mathrm{X}_i)$ and standard deviation $\sigma_Y(\mathrm{X}_i)$, and then decoded back assuming unimodal normally distributed noise?:

$$\widetilde{\varphi_{Y|X}}(Y_j|\mathrm{X}_i) = N(Y_j; \mu_Y(\mathrm{X}_i), \sigma_Y(\mathrm{X}_i)) \quad (1)$$

where $\mu_Y(\mathrm{X}_i) = \sum_{j=1}^{k} Y_j \cdot \varphi_{Y|X}(Y_j|X_i)$ and $\sigma_Y^2(\mathrm{X}_i) = \sum_{j=1}^{k}(Y_j - \mu_Y(\mathrm{X}_i))^2 \cdot \varphi_{Y|X}(Y_j|X_i)$.

In the few cases when $\sigma_Y(\mathrm{X}_i)$=0, the distribution (1) has been replaced by Dirac delta distribution spiking at the grid point nearest to $\mu_Y(\mathrm{X}_i)$ position.

Unimodal distributions can be seen as a representation of dual response: the signal – i.e., the mode, and the noise – i.e., the dispersion around it. The dual response representation of stochastic outcomes is a valuable technique that is widely used in time series analysis addressing heteroskedasticity (e.g., GARCH models), design of experiments (Vining and Myers, 1990; Castillo and Montgomery, 1993; Lin and Tu, 1995), and robust tuning of machine learning models (Glushkovsky, 2018).

For comparison, a single response represents the deterministic case, which is out of scope of this paper, while a triple response represents more complex encoding addressing asymmetry of conditional distributions in addition to the mean and standard deviation. A triple response requires estimation of the first three moments and then fitting the empirical conditional distributions that often leads to overfitting considering the complexity of the approach and limited degree of freedom.

Notably, the encoding of the input data into mean and standard deviation is also the key concept of AI generative models, particularly variational autoencoders (VAE), followed by decoding that reconstructs the data (Kingma and Welling, 2014).

Combining decoded conditional distributions (1) across all $X_i$, $i \in [1, k]$, it is possible to compose the anticipated 2D joint distribution assuming $X \rightarrow Y$ cause-effect relationship:

$$\widetilde{\widetilde{\varphi_{X \rightarrow Y}}}(X_i, Y_j) = \frac{\widetilde{\varphi_{Y|X}}(Y_j|\mathrm{X}_i)}{\sum_{j=1}^{k} \widetilde{\varphi_{Y|X}}(Y_j|\mathrm{X}_i)} \cdot \sum_{j=1}^{k} \varphi(X_i, Y_j), i, j \in [1, k] \quad (2)$$

where the denominator has been introduced to normalize the decoded empirical distribution.

The comparison between the original $\varphi(X_i, Y_j)$ and the decoded $\widetilde{\widetilde{\varphi_{X \rightarrow Y}}}(X_i, Y_j)$ distributions indicates how much information is lost after encoding into the "bottleneck" of the dual response, i.e., the mean $\mu_Y(\mathrm{X}_i)$ and the gaussian noise N(0, $\sigma_Y(\mathrm{X}_i)$) around it. The closer the distributions, the less information is lost representing effect *Y* by dual response. On the contrary, the higher the deviation between the original and the decoded distributions, the more factors exist in the assumed effect *Y* in addition to the anticipated one that was caused by *X*. The former supports the assumption that there is a $X \rightarrow Y$ cause-effect relationship, while the latter points to the opposite.

Similarly, assuming a $Y \rightarrow X$ cause-effect relationship, it is possible to observe deviation between the original $\varphi(X_i, Y_j)$ and the decoded $\widetilde{\widetilde{\varphi_{Y \rightarrow X}}}(X_i, Y_j)$ distributions.

Examples of the original (a) and anticipated asymmetric (b and c) 2D probability density distributions are shown in Figure 1.

It can be observed that the original 2D probability density distribution with true $X \rightarrow Y$ cause-effect relationship in Figure 1, (a) is closer to the anticipated distribution in Figure 1 (b) than in Figure 1 (c). This observation supports correct identification of causal directionality.

Considering the results of both $X \rightarrow Y$ and $Y \rightarrow X$ assumptions, it supports the decision on causal direction:

$$\begin{cases} \text{If } \mathbf{M}\left[\varphi(X_i, Y_j), \widetilde{\widetilde{\varphi_{Y \rightarrow X}}}(X_i, Y_j)\right] > \mathbf{M}\left[\varphi(X_i, Y_j), \widetilde{\widetilde{\varphi_{X \rightarrow Y}}}(X_i, Y_j)\right], \textit{then } \mathrm{X} \rightarrow \mathrm{Y} \\ \text{If } \mathbf{M}\left[\varphi(X_i, Y_j), \widetilde{\widetilde{\varphi_{Y \rightarrow X}}}(X_i, Y_j)\right] < \mathbf{M}\left[\varphi(X_i, Y_j), \widetilde{\widetilde{\varphi_{X \rightarrow Y}}}(X_i, Y_j)\right], \textit{then } \mathrm{Y} \rightarrow \mathrm{X} \end{cases} \quad (3)$$

where the deviation metric $\mathbf{M}[\cdot,\cdot]$ between distributions follows the rule without loss of generality: "the lower the value of the metric, the closer the distributions are".

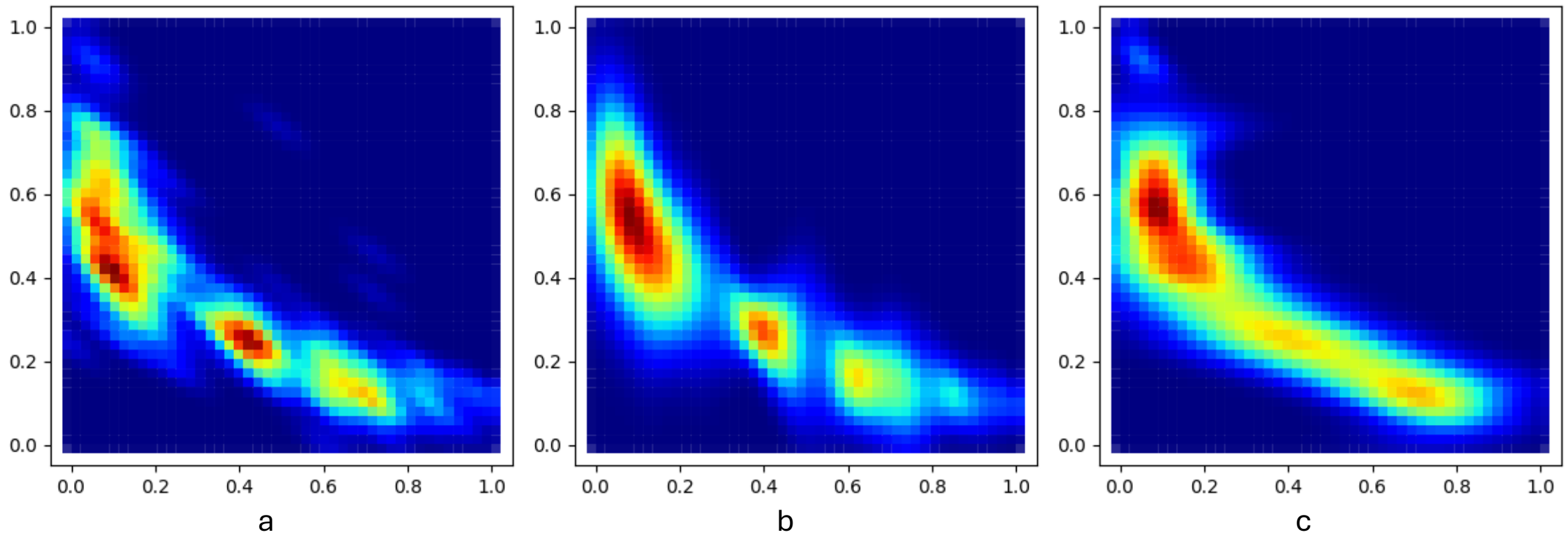

Figure 1. Examples of 2D probability density distributions for ID = 13:

(a) original $\varphi(X_i, Y_j)$ with true $X \rightarrow Y$ cause-effect relationship
(b) anticipated $\widetilde{\widetilde{\varphi_{X\rightarrow Y}}}(X_i, Y_j)$ assuming $X \rightarrow Y$ cause-effect relationship
(c) anticipated $\widetilde{\widetilde{\varphi_{Y\rightarrow X}}}(X_i, Y_j)$ assuming $Y \rightarrow X$ cause-effect relationship

In practice, inequalities (3) rarely reduce to an equality. This transition occurs exclusively when both anticipated distributions are identical having $\widetilde{\widetilde{\varphi_{Y\rightarrow X}}}(X_i, Y_j) = \widetilde{\widetilde{\varphi_{X\rightarrow Y}}}(X_i, Y_j)$ across all $(X_i, Y_j)$ values. Notably, these indecisive cases occurred in only two real-world pairs (IDs 44 and 84). These pairs feature highly correlated data (99.9% and 99.5%, respectively) and only produced these results when using coarse grids with $k$ set to 15 or 20.

The deviation metric can be estimated by applying different approaches, such as correlation, cosine distance, entropy (i.e., the difference in information content), Jaccard index, Kullback–Leibler (K-L) divergence, Kolmogorov-Smirnov (K-S) distance, mean absolute error (MAE), mean squared error (MSE), mutual information, or Wasserstein distance. More details regarding the applied metrics and the obtained results are presented in Sections 6 and 8.

Diagram of the causal directionality identification applying AAG is illustrated in Figure 2.

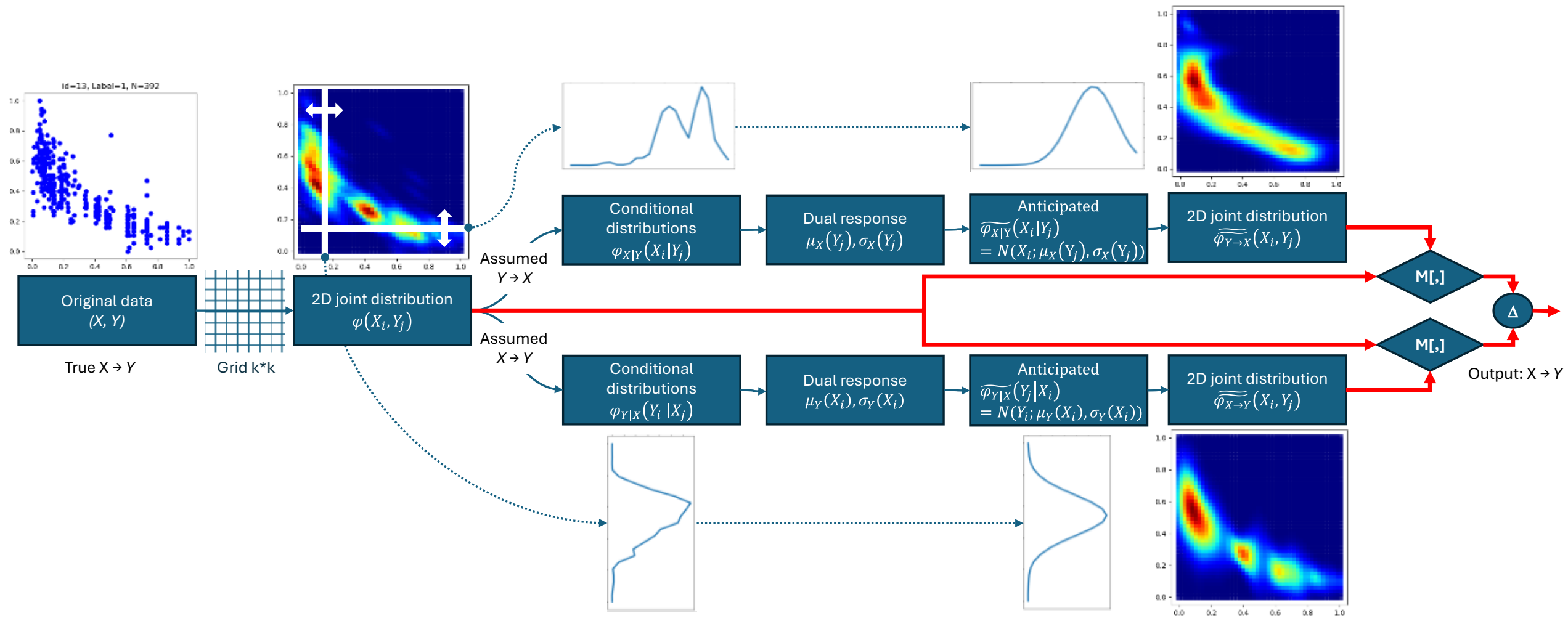

Figure 2. Diagram of causal directionality identification applying AAG method

## 4. Monotonicity Index Methods

Modality can be expressed as the number of times the gradient of the probability density function changes its' sign along an axis. Practically, the *numpy.gradient* has been used to calculate gradients.

Example of gradient vectors is shown in Figure 3.

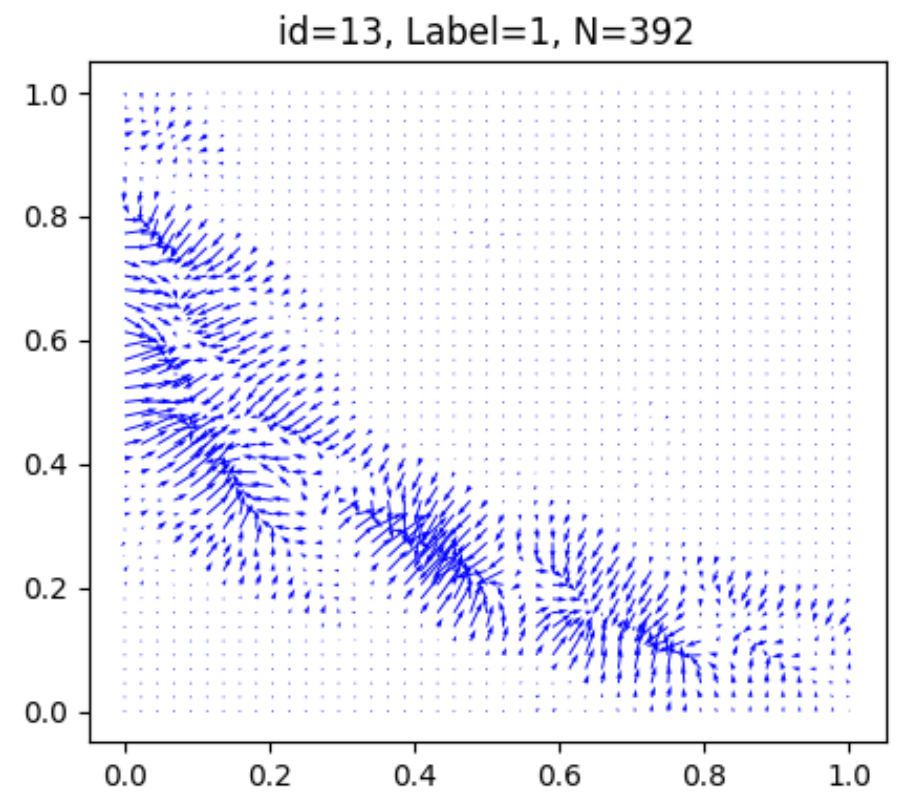

Figure 3. Example of gradient vectors of $\varphi(X_i, Y_j)$ for ID = 13

Monotonicity indexes are calculated as weighted sums of counts of gradient sign changes along both $X$ and $Y$ axes of the 2D joint probability distribution function $\varphi(X_i, Y_j)$:

$$\begin{cases} MI_X = \sum_{i=1}^{k-1} \sum_{j=1}^{k} \omega_X(X_i, Y_j) \cdot \mathbf{1}\left(\left| sgn\left(\nabla_X\left(\varphi(X_{i+1}, Y_j)\right)\right) - sgn\left(\nabla_X\left(\varphi(X_i, Y_j)\right)\right)\right|\right) \\ MI_Y = \sum_{i=1}^{k} \sum_{j=1}^{k-1} \omega_Y(X_i, Y_j) \cdot \mathbf{1}\left(\left| sgn\left(\nabla_Y\left(\varphi(X_i, Y_{j+1})\right)\right) - sgn\left(\nabla_Y\left(\varphi(X_i, Y_j)\right)\right)\right|\right) \end{cases} \quad (4)$$

where $\mathbf{1}(\cdot)$ is the unit step function, $\nabla_X(\cdot)$ and $\nabla_Y(\cdot)$ are gradients along $X$ and $Y$ axes, consequently, $sgn(\cdot)$ is the sign function, and $\omega_X$, $\omega_Y$ are weights reducing the impacts of instabilities of distributions gradients, especially on tails.

Weights provide regularization that boosts robustness of the approach. They diminish noisy non-monotonic effects considering both marginal distributions and absolute values of gradients. Thus, noisy effects of the oscillating gradients on tails of distributions have been diminished by the low values of the marginal distributions:

$$\begin{cases} \varphi_X(X_i) = \sum_{j=1}^{k} \varphi(X_i, Y_j) \\ \varphi_Y(Y_j) = \sum_{i=1}^{k} \varphi(X_i, Y_j) \end{cases} \quad (5)$$

while noisy effects of the oscillating gradients along entire distributions have been reduced in case of low amplitudes of the gradients:

$$\begin{cases} \gamma_X(X_i, Y_j) = 0.5 \cdot \left(\left|\nabla_X\left(\varphi(X_{i+1}, Y_j)\right)\right| + \left|\nabla_X\left(\varphi(X_i, Y_j)\right)\right|\right) \\ \gamma_Y(X_i, Y_j) = 0.5 \cdot \left(\left|\nabla_Y\left(\varphi(X_i, Y_{j+1})\right)\right| + \left|\nabla_Y\left(\varphi(X_i, Y_j)\right)\right|\right) \end{cases} \quad (6)$$

The factors (6) take into consideration the averages of absolute values of gradients of neighboring points involved in the monotonicity indexes calculations (4).

Two types of regularizations have been considered:

a) simple products between (5) and (6): $\omega_X = \varphi_X \cdot \gamma_X$, and $\omega_Y = \varphi_Y \cdot \gamma_Y$, correspondingly
b) insensitive zone $\gamma^*$ applied to factors (6): $\omega_X = \varphi_X \cdot \mathbf{1}(\gamma_X - \gamma^*)$, and $\omega_Y = \varphi_Y \cdot \mathbf{1}(\gamma_Y - \gamma^*)$.

The lower the monotonicity index, the closer the distribution to a unimodal one. Comparison of two indexes (4) supports the decision on causal direction:

$$\begin{cases} \text{If } MI_X > MI_Y, then \text{ X } \rightarrow \text{ Y} \\ \text{If } MI_X < MI_Y, then \text{ Y } \rightarrow \text{ X} \end{cases} \quad (7)$$

## 5. Data Set and Data Preparation

Bivariate numerical data used in this article are the Tübingen real-world examples from (Mooij *et al*, 2014). It allows for benchmarking of the proposed methods to Additive Noise Models approach (Hoyer *et al*, 2009; Mooij *et al*, 2016), Generalized Root Causal Inference (GRCI) (Strobl and Lasko, 2023), and location-scale noise models with independence testing (CAREFL-H) (Xiangyu Sun and Schulte, 2023).

Subset of 102 pairs out of 108 have been selected keeping only two-dimensional cases, while pairs number 52-55, 71 and 105 have been excluded.

The normalization $X, Y \in [0,1]$ has been applied across all pairs for simplicity without loss of generality.

## 6. Applied Metrics

Seventeen alternative metrics **M** have been applied to measure deviation between the original and the anticipated distributions (3):

- cosine distance (COS), i.e., dissimilarity, based on dot product:
  $$\mathbf{M}[\varphi(X_i,Y_j),\tilde{\tilde{\varphi}}(X_i,Y_j)] = 1 - \frac{\sum_{i=1}^{k}\sum_{j=1}^{k}\varphi(X_i,Y_j)\cdot\tilde{\tilde{\varphi}}(X_i,Y_j)}{||\varphi(X_i,Y_j)||\cdot||\tilde{\tilde{\varphi}}(X_i,Y_j)||}$$
  Practically, *cosine* from *scipy.spatial.distance* has been used
- entropy, i.e., the difference in information content:
  $$\mathbf{M}[\varphi(X_i,Y_j),\tilde{\tilde{\varphi}}(X_i,Y_j)] = \sum_{i=1}^{k}\sum_{j=1}^{k}\left(-\varphi(X_i,Y_j)\cdot\log\left(\varphi(X_i,Y_j)\right) + \tilde{\tilde{\varphi}}(X_i,Y_j)\cdot\log\left(\tilde{\tilde{\varphi}}(X_i,Y_j)\right)\right)$$
- Hellinger distance (Hellinger, 1909):
  $$\mathbf{M}[\varphi(X_i,Y_j),\tilde{\tilde{\varphi}}(X_i,Y_j)] = \frac{1}{\sqrt{2}}\sqrt{\sum_{i=1}^{k}\sum_{j=1}^{k}\left(\sqrt{\varphi(X_i,Y_j)} - \sqrt{\tilde{\tilde{\varphi}}(X_i,Y_j)}\right)^2}$$
- Jeffreys divergence, also known as information value (Jeffreys, 1946).:
  $$\mathbf{M}[\varphi(X_i,Y_j),\tilde{\tilde{\varphi}}(X_i,Y_j)] = \sum_{i=1}^{k}\sum_{j=1}^{k}\left(\varphi(X_i,Y_j) - \tilde{\tilde{\varphi}}(X_i,Y_j)\right)\cdot\log(\frac{\varphi(X_i,Y_j)}{\tilde{\tilde{\varphi}}(X_i,Y_j)})$$
- Jensen–Shannon divergence, i.e., the symmetrical K-L metric (Lin, 1991):
  $$\mathbf{M}[\varphi(X_i,Y_j),\tilde{\tilde{\varphi}}(X_i,Y_j)] = \frac{1}{2}\sum_{i=1}^{k}\sum_{j=1}^{k}(\varphi(X_i,Y_j)\cdot\log(\frac{\varphi(X_i,Y_j)}{\bar{\bar{\varphi}}(X_i,Y_j)}) + \tilde{\tilde{\varphi}}(X_i,Y_j)\cdot\log(\frac{\tilde{\tilde{\varphi}}(X_i,Y_j)}{\bar{\bar{\varphi}}(X_i,Y_j)}))$$
  where $\bar{\bar{\varphi}}(X_i,Y_j) = 0.5\cdot(\varphi(X_i,Y_j) + \tilde{\tilde{\varphi}}(X_i,Y_j))$
- Kullback–Leibler (K-L) divergence (Kullback and Leibler, 1951):
  $$\mathbf{M}[\varphi(X_i,Y_j),\tilde{\tilde{\varphi}}(X_i,Y_j)] = \sum_{i=1}^{k}\sum_{j=1}^{k}\varphi(X_i,Y_j)\cdot\log(\frac{\varphi(X_i,Y_j)}{\tilde{\tilde{\varphi}}(X_i,Y_j)})$$
  where $\tilde{\tilde{\varphi}}$ is either $\widetilde{\widetilde{\varphi_{X\to Y}}}$ or $\widetilde{\widetilde{\varphi_{Y\to X}}}$, and if $\varphi = 0$, then $\log(\varphi) = 0$
- maximum absolute error (MaxAE):
  $\mathbf{M}[\varphi(X_i,Y_j),\tilde{\tilde{\varphi}}(X_i,Y_j)] = max_{i,j}\left|\varphi(X_i,Y_j) - \tilde{\tilde{\varphi}}(X_i,Y_j)\right|, i,j \in [1,k]$
- maximum density: $max_{i,j}(\widetilde{\widetilde{\varphi_{Y\to X}}}(X_i,Y_j)) - max_{i,j}(\widetilde{\widetilde{\varphi_{X\to Y}}}(X_i,Y_j))$
- mean absolute error (MAE):
  $$\mathbf{M}[\varphi(X_i,Y_j),\tilde{\tilde{\varphi}}(X_i,Y_j)] = \frac{1}{k^2}\sum_{i=1}^{k}\sum_{j=1}^{k}\left|\varphi(X_i,Y_j) - \tilde{\tilde{\varphi}}(X_i,Y_j)\right|$$

- mean squared error (MSE):

$$\mathbf{M}\left[\varphi(X_i,Y_j),\tilde{\tilde{\varphi}}(X_i,Y_j)\right]=\frac{1}{k^2}\sum_{i=1}^{k}\sum_{j=1}^{k}\left(\varphi(X_i,Y_j)-\tilde{\tilde{\varphi}}(X_i,Y_j)\right)^2$$

- mutual information:
  - (a) Generally scaled *mutual_info_score* from *sklearn.metrics* (MI)
  - (b) Normalized *normalized_mutual_info_score* from *sklearn.metrics* (MI_norm)
- Pearson correlation
- 2D Wasserstein distance that measures the minimum "cost" of turning one probability distribution into another (Villani, 2009)
- 2D Kolmogorov-Smirnov maximum distance (2D K-S Max) (Peacock, 1983):

$$\mathbf{M}\left[\varphi(X_i,Y_j),\tilde{\tilde{\varphi}}(X_i,Y_j)\right]=max_{i,j}\left|F(X_i,Y_j)-\tilde{\tilde{F}}(X_i,Y_j)\right|$$

  where $F(X_i,Y_j)=\sum_{k=1}^{i}\sum_{l=1}^{j}\varphi(x_k\leq X_i,y_l\leq Y_j)$ is 2D cumulative distribution function
- 2D root means squared error between two cumulative distributions that is an equivalent to discrete Cramér–von Mises discrepancy (2D CDF RMSE) (Choulakian, 1994):

$$\mathbf{M}\left[\varphi(X_i,Y_j),\tilde{\tilde{\varphi}}(X_i,Y_j)\right]=\sqrt{\frac{1}{k^2}\sum_{i=1}^{k}\sum_{j=1}^{k}\left(F(X_i,Y_j)-\tilde{\tilde{F}}(X_i,Y_j)\right)^2}$$

- Jaccard index that equals the ratio of intersection to union areas of original and anticipated contours of distributions. Nine areas have been defined considering Lebesque levels of the probability density values:

$$\delta_m(X_i,Y_j)=\begin{cases}1,\text{ if }\varphi(X_i,Y_j)>m\cdot max_{i,j}\left(\varphi(X_i,Y_j)\right)/10, m\in[1,9]\\0,\text{ otherwise}\end{cases}$$

Examples of contour plots of probability density distributions are shown in Figure 4, where dark plum lines represent $m$=1 and yellow lines represent $m$=9. It can be observed that the Jaccard index has a higher value of 0.48 for areas displayed on chart (d) compared to 0.37 on chart (e) and, therefore, it correctly suggests a $X\rightarrow Y$ cause-effect relationship.

## 7. Hyperparameters and Applied Tuning

The described AAG and Monotonicity Index methods include only two or three hyperparameters that affect the accuracy of the causal directionality identifications. However, for a given set of hyperparameters, either AAG or Monotonicity Index methods provide a unique deterministic outcome of the solution. Methods do not contain stochastic learning processes, such as gradient decent in neural network trainings that require defined initial values of weights. It means complete repeatability of the results without concerns of solutions convergence.

Acknowledging that the results are sensitive to hyperparameters, tuning of hyperparameters has been done by applying full factorial Design of Experiment (DOE) (Table 1), where $k$ is a grid size, *bw_par* is a scalar of *stats.gaussian_kde* method of fitting empirical 2D joint probability density distribution, $m$ is a level of contour (Figure 4), and $\gamma^*$ is the insensitive zone of the regularization factor of the Monotonicity Index method. The first two are general hyperparameters across both methods, while the last two are dedicated to a specific metric or method.

| | Tuning Design of Experiment | | |
|---|---|---|---|
| Hyperparameter | # of Levels | Min | Max |
| *k* | 10 | 15 | 65 |
| *bw_par* | 10 | 0.025 | 0.250 |
| *m* | 5 | 5 | 9 |
| $\gamma^*$ | 5 | 1.00E-11 | 1.00E-07 |

Table 1. Factors (hyperparameters) of full factorial Design of Experiment

It should be noted that applied levels of hyperparameters have significant ranges aiming to find global optimization not a local maximum.

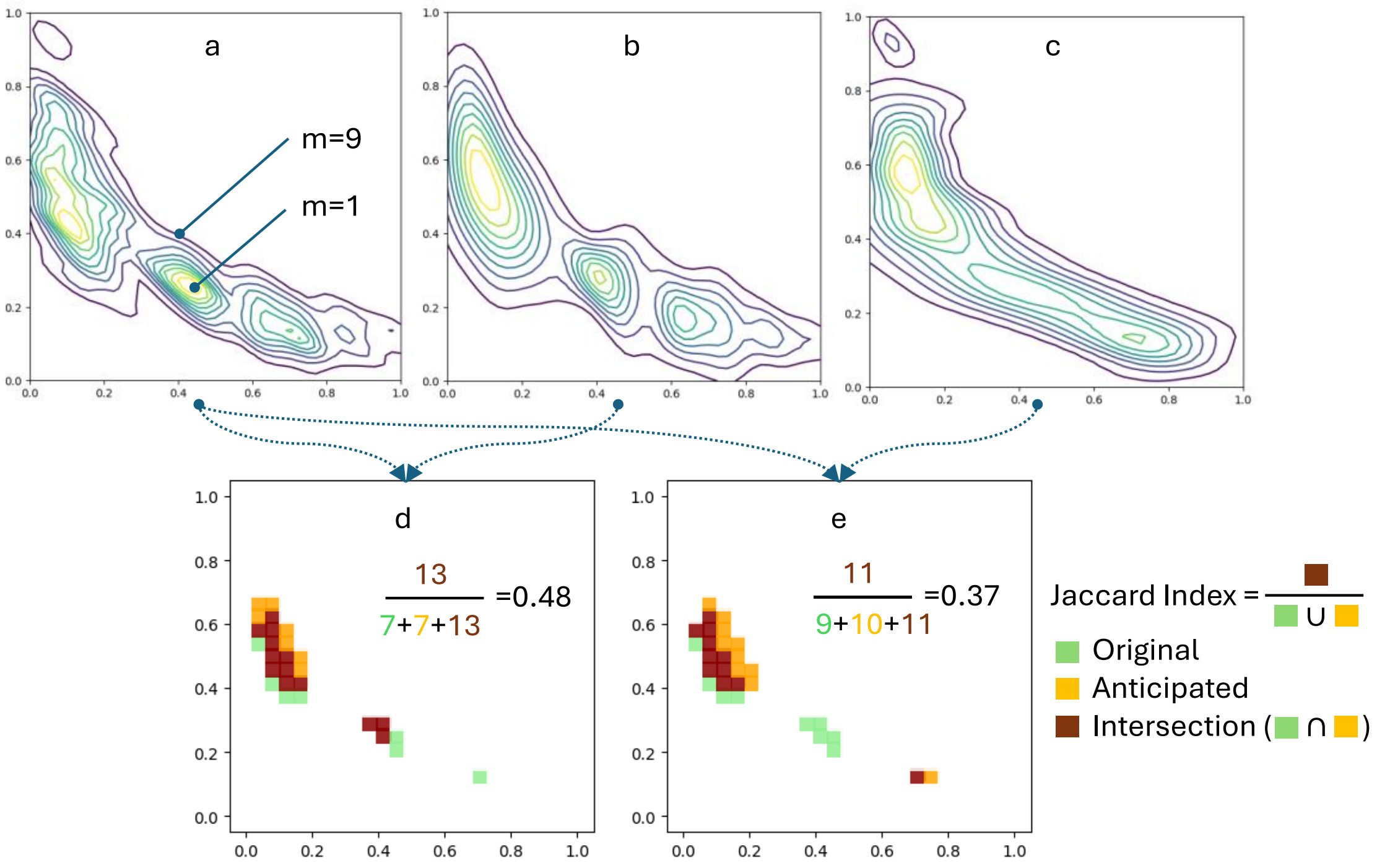

Figure 4. Examples of contour plots of probability density distributions (ID = 13):

(a) original with true $X \longrightarrow Y$ cause-effect relationship
(b) anticipated assuming $X \longrightarrow Y$ cause-effect relationship
(c) anticipated assuming $Y \longrightarrow X$ cause-effect relationship

and intersected and union areas for $m$=7:

(d) between original (a) and anticipated assuming $X \longrightarrow Y$ cause-effect relationship (b)
(e) between original (a) and anticipated assuming $Y \longrightarrow X$ cause-effect relationship (c)

Two types of tuning have been considered: simple and size-adaptive. The former aims to simply tune hyperparameters that maximize accuracy. The latter considers interactive effects between hyperparameters and sample sizes of pairs (Section 9).

## 8. Identifications of Causal Directionality

The weighted accuracies of causal directionality identification following simple tunings of hyperparameters are presented in Table 2 by applying two methods: (1) AAG with seventeen different metrics **M** and (2) Monotonicity Index with two alternative regularizations. Additionally, the table presents the values of the tuned hyperparameters after running Design of Experiment according to Table 1.

The suggested weighting of accuracy rate has been applied to mitigate potential bias of the data sources (Mooij *et al*, 2016; Strobl and Lasko, 2023). Notably, the weighting approach influenced the obtained results. The cosine distance metric yielded fewer misclassifications —24 out of 102 compared to the 27 pairs misclassified by the 2D CDF RMSE metric. Nevertheless, the total weight of the 24 pairs was higher than that of the 27 pairs (8.12 and 7.70, respectively).

The tuned versions of the top 12 AAG methods (out of 17) outperformed the Monotonicity Index. The highest performance was achieved using metrics: 2D CDF RMSE, MSE, and cosine distance, reaching a top weighted accuracy of 78.6% classifying 102 pairs of the Tübingen real-world examples from (Mooij *et al*, 2014). As expected, Pearson correlation and cosine distance, provided close results due to preceding data normalization.

| | | | Tuned Hyperparameters | | | |
|---|---|---|---|---|---|---|
| Method | Metric | Weighted Accuracy | *k* | *bw_par* | *m* | *γ** |
| AAG | 2D CDF RMSE | 78.6% | 30 | 0.125 | | |
| AAG | MSE | 77.7% | 25 | 0.175 | | |
| AAG | Cosine Distance | 77.4% | 25 | 0.175 | | |
| AAG | Pearson Correlation | 77.4% | 25 | 0.175 | | |
| AAG | 2D K-S Max | 77.1% | 60 | 0.050 | | |
| AAG | 2D Wasserstein | 75.8% | 30 | 0.050 | | |
| AAG | K-L Divergence | 75.7% | 20 | 0.050 | | |
| AAG | MAE | 75.2% | 30 | 0.100 | | |
| AAG | MaxAE | 73.6% | 15 | 0.100 | | |
| AAG | Hellinger Distance | 73.3% | 30 | 0.100 | | |
| AAG | J-S Divergence | 72.9% | 30 | 0.100 | | |
| AAG | Jaccard Index | 72.5% | 55 | 0.025 | 8 | |
| AAG | Jeffreys Divergence | 67.9% | 55 | 0.075 | | |
| AAG | Mutual Information | 66.7% | 20 | 0.075 | | |
| AAG | Maximum Density | 65.6% | 15 | 0.075 | | |
| AAG | Entropy | 65.6% | 20 | 0.025 | | |
| AAG | Mutual Inform Normalized | 65.6% | 20 | 0.075 | | |
| Monot Index | Weighted | 70.6% | 45 | 0.150 | | 1.50E-11 |
| Monot Index | Insensitive Zone | 70.6% | 45 | 0.150 | | |

Table 2. Weighted top accuracy results with corresponding simply tuned hyperparameters (102 pairs)

## 9. Impact of Sample Sizes and Size-Adaptive Tuning

The dataset of 102 pairs has a wide range of sample sizes varying between 94 and 16,382 observations. It is reasonable to assume sensitivity of AAG method to the number of observations. Specifically, the approximation of empirical 2D probability density function $\varphi(X_i, Y_j)$, that is common across all described methods, depends on interactive effects between sample size and hyperparameters ($k$ and $bw_par$). Practically, it suggests the size-adaptive tuning of the hyperparameters. Study of this approach reveals that the size-adaptive tuning of two hyperparameters can substantially improve accuracy of the method.

For example, size-adaptive tuning, which is based on a single binary split of pairs with sample sizes below or above 350 observations, boosts accuracy by about +2.2%. Results of size-adaptive tuning of hyperparameters are shown in Table 3.

| | | Size-adaptive Tuning of Hyperparameters | | |
|---|---|---|---|---|
| Metric | Weighted Accuracy | Split point by # of obs | *k* | *bw_par* |
| 2D CDF RMSE | 80.1% | <= 350 | 30 | 0.050 |
| | | > 350 | 30 | 0.125 |
| MSE | 80.4% | <= 350 | 30 | 0.050 |
| | | > 350 | 25 | 0.175 |
| Cosine Distance | 79.7% | <= 350 | 30 | 0.050 |
| | | > 350 | 25 | 0.175 |

Table 3. AAG optimized size-adaptive tunings of hyperparameters (102 pairs)

Binning into only two intervals has been dictated by quite a limited number of pairs in the dataset and by risk of overfitting. The selected split point of 350 observations provides quite balanced distribution of pairs: 42 to be below the split point and 60, otherwise.

Notably, the size-adaptive tuning of hyperparameters for pairs with the number of observations above 350 mostly aligns with the simple tuning, while pairs with smaller sample sizes require more granular grid and/or less smoothed approximation of the 2D probability density function.

The provided example of the size-adaptive tuning is rather directional than conclusive and requires further studies including more real-world data (see Section 12).

## 10. Benchmarking of Results

Identifications of casual directionality by tuned top AAG methods against benchmark models are shown in Table 4. It includes results of applying both single and size-adaptive tuning approaches and different exclusions of pairs to have comparable data.

| | | Weighted Accuracy | | | | | |
|---|---|---|---|---|---|---|---|
| **Method** | **Tuning** | **2D CDF RMSE** | **MSE** | **COS** | **Number of pairs** | **Excluded pairs** | **Reasons of exclusion** |
| AAG | Size-adaptive | 80.1% | 80.4% | 79.7% | 102 | 52-55, 71, 105 | Multivariate |
| AAG | Simple | 78.6% | 77.7% | 77.4% | 102 | 52-55, 71, 105 | Multivariate |
| AAG | Size-adaptive | 86.1% | 86.4% | 85.7% | 99 | 17, 52-55, 71, 87, 105, 107 | Multivariate; MI < 0.125 |
| AAG | Simple | 84.5% | 83.5% | 83.2% | 99 | 17, 52-55, 71, 87, 105, 107 | Multivariate; MI < 0.125 |
| AAG | Size-adaptive | 84.4% | 84.7% | 84.0% | 99 | 47, 52-55, 70, 71, 105, 107 | Multivariate; binary |
| AAG | Simple | 82.7% | 81.7% | 81.4% | 99 | 47, 52-55, 70, 71, 105, 107 | Multivariate; binary |
| CAREFL-H | | 82.0% | | | 99 | 47, 52-55, 70, 71, 105, 107 | Multivariate; binary |
| GRCI | | 81.6% | | | 99 | 47, 52-55, 70, 71, 105, 107 | Multivariate; binary |
| AAG | Size-adaptive | 88.4% | 88.8% | 88.0% | 97 | 17, 47, 52-55, 70, 71, 87, 105, 107 | Multivariate; binary; MI < 0.125 |
| AAG | Simple | 86.7% | 85.6% | 85.3% | 97 | 17, 47, 52-55, 70, 71, 87, 105, 107 | Multivariate; binary; MI < 0.125 |
| AAG | Size-adaptive | 82.4% | 82.4% | 81.6% | 95 | 52-55, 71, 101-108 | Multivariate; IDs > 100 |
| AAG | Simple | 80.8% | 80.2% | 79.9% | 95 | 52-55, 71, 101-108 | Multivariate; IDs > 100 |
| ANM | | 63 ± 10 % | | | 95 | 52-55, 71, 101-108 | Multivariate; IDs > 100 |

Table 4. Weighted accuracy results of tuned top AAG methods against benchmark approaches

To benchmark the obtained results to ANM that achieved a weighted accuracy of 63 ± 10 % and an AUC of 0.74 ± 0.05 (Hoyer *et al*, 2009), pairs with numbers higher than 100 have been removed to match the data of 95 pairs. For that population, the weighted accuracy of 80.8% and AUC of 0.83 (Figure 5) have been reached when applying the AAG method with 2D CDF RMSE metric **M** and simple tuning of hyperparameters ($k$=30 and $bw_par$=0.125). The increased accuracy of 82.4% has been achieved by applying size-adaptive tuning of hyperparameters.

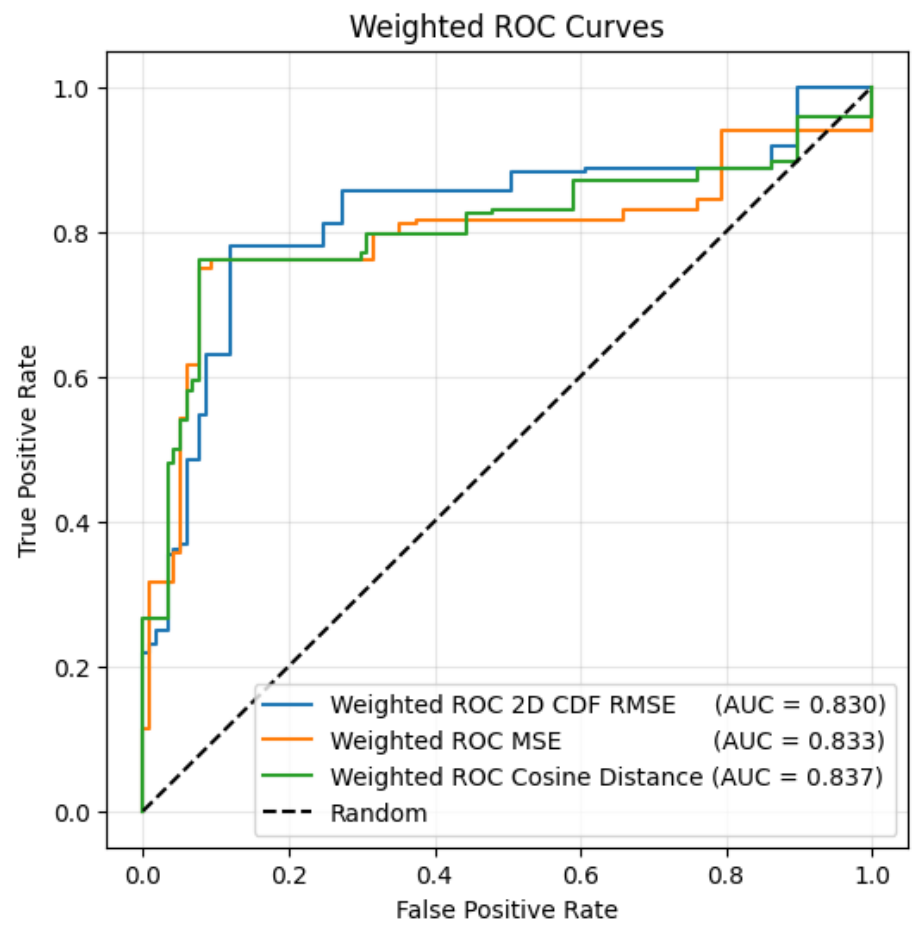

Figure 5. ROC charts applying tuned AAG methods (95 pairs)

To compare the AAG method against GRCI (Strobl and Lasko, 2023) and CAREFL-H (Xiangyu Sun and Schulte, 2023) approaches, tree pair numbers 47, 70, and 107 with binary variables have been excluded out of 102. Based on remaining 99 pairs, the AAG method based on the 2D CDF RMSE metric reaches weighted accuracy of 82.7% or 84.4% using simple or size-adaptive tuning, correspondingly. It shows outperforming results versus 81.6% obtained by GRCI or 82% achieved by CAREFL-H.

Notably, the AAG method is a straightforward approach: once the empirical 2D probability density function is fitted, the method produces a deterministic outcome without requiring additional model optimization or convergence of an iterative process.

## 11. Robustness of the Obtained Results

Tuning optimization is dictated by data with known results of causal directionality. It means that for other datasets, the identification results of causal directionality using tuned hyperparameters shown in (Table 2 and 3) may be suboptimal, i.e., there is a potential risk of hyperparameter's overfitting.

To demonstrate consistency and robustness of the described AAG and Monotonicity Index methods across wide ranges of hyperparameter's combinations, results of the performed DOE (Table 1) have been compiled in Table 5 presenting statistics of the weighted accuracies, such as medians, lower (5%) and upper (95%) confidence limits, and confidence intervals.

| | | Weighted Accuracy | | | |
|---|---|---|---|---|---|
| **Method** | **Metric** | **LCL_5%** | **Median** | **UCL_95%** | **CL_Interval** |
| AAG | 2D CDF RMSE | 66.2% | 74.2% | 78.5% | 12.2% |
| AAG | MSE | 61.0% | 66.9% | 73.3% | 12.3% |
| AAG | Cosine Distance | 60.5% | 66.9% | 73.0% | 12.5% |
| AAG | 2D K-S Max | 62.3% | 69.1% | 72.8% | 10.5% |
| AAG | MAE | 61.9% | 67.5% | 72.8% | 10.9% |
| AAG | Pearson Correlation | 59.9% | 66.8% | 72.6% | 12.7% |
| AAG | 2D Wasserstein | 63.5% | 67.7% | 72.5% | 9.0% |
| AAG | J-S Divergence | 61.0% | 66.0% | 69.8% | 8.7% |
| AAG | MaxAE | 57.1% | 64.2% | 69.6% | 12.5% |
| AAG | Hellinger Distance | 61.3% | 65.3% | 69.1% | 7.8% |
| AAG | K-L Divergence | 60.2% | 63.0% | 68.3% | 8.1% |
| AAG | Jaccard Index | 54.1% | 60.7% | 67.6% | 13.6% |
| AAG | Jeffreys Divergence | 55.5% | 63.6% | 67.2% | 11.6% |
| AAG | Maximum Density | 49.7% | 55.8% | 64.0% | 14.3% |
| AAG | Mutual Information | 33.1% | 46.2% | 63.8% | 30.8% |
| AAG | Entropy | 40.0% | 47.6% | 59.7% | 19.7% |
| AAG | Mutual Inform Normalized | 33.2% | 44.1% | 57.0% | 23.9% |
| Monot Index | Weighted | 53.9% | 63.7% | 70.6% | 16.7% |
| Monot Index | Insensitive Zone | 54.9% | 62.7% | 68.6% | 13.7% |

Table 5. Weighted accuracy results across all combinations of DOE hyperparameters (102 pairs)

According to the upper confidence limits, the top three performing methods are AAG with 2D CDF RMSE, MSE, and cosine distance metrics. The close confidence intervals across these top three methods, along with the reasonable accuracy drops from the tuned runs (Table 2) to the upper confidence limits, demonstrates the robustness of these top three approaches regarding their hyperparameters.

To visualize the sensitivity issue, accuracy mapping across full factorial DOE of hyperparameters *k* and *bw_par* by applying AAG with top three metrics have been explored (Figure 6).

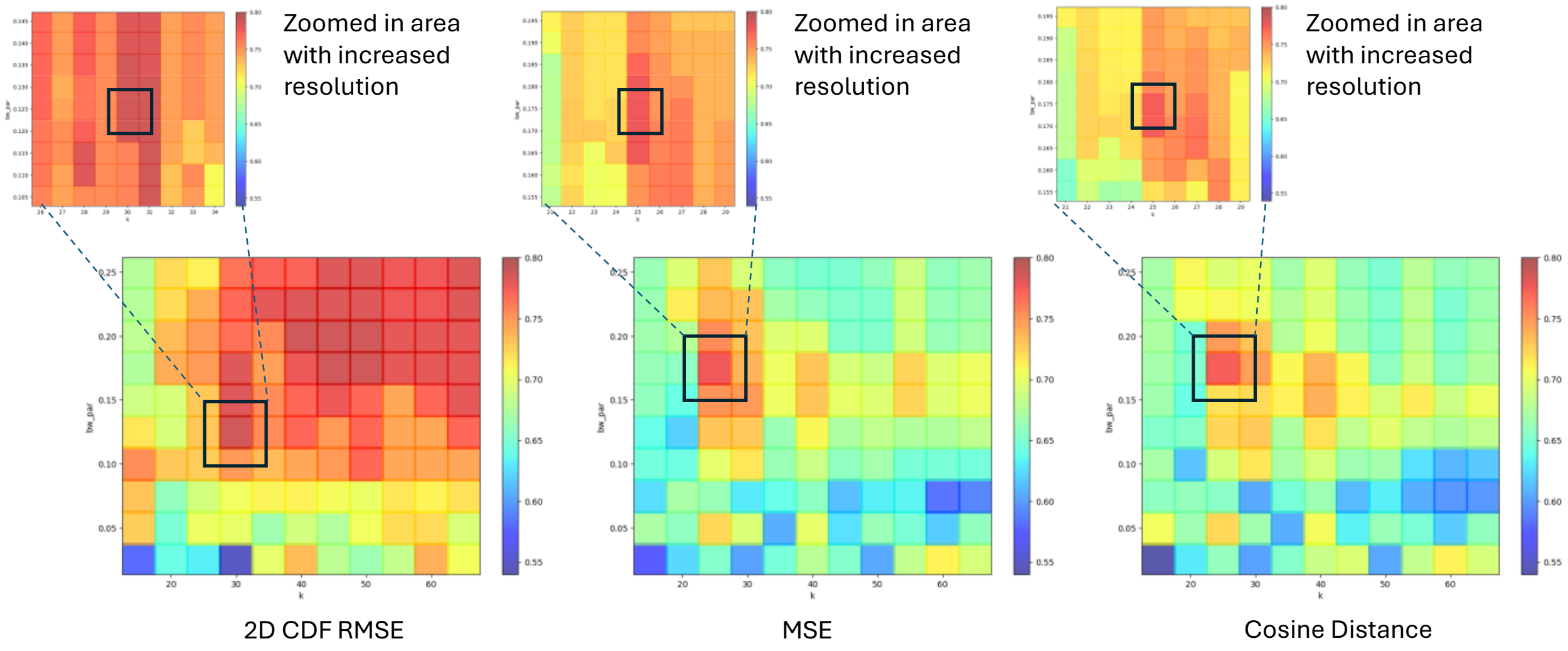

Figure 6. Weighted accuracy mappings across full factorial DOE of hyperparameters '*k*' and '*bw_par*' by applying AAG methods (102 pairs)

The maps demonstrate a relatively smooth accuracy surface around the maximization points for MSE and cosine distance metrics, and this is particularly evident for the 2D CDF RMSE. To examine the results further, the regions surrounding the maximum points were analyzed at higher hyperparameter resolutions. The stability of the maximum points across all three metrics provides evidence of robust hyperparameter tuning. Overall, these results suggest that the AAG approach is reasonably robust with respect to hyperparameter sensitivity.

Furthermore, to address out-of-sample generalization that may expose overfitting, 5-times repeated stratified 10-fold cross-validation with hyperparameter tuning has been performed. This approach is a very common in machine learning having very limited number of observations, while the classical approach of controlling overfitting by splitting data into only two train and test subsets is not applicable given a limited number of pairs. The results of cross-validation, featuring preceding per-repeat summarizations, are presented in Table 6.

| | Weighted Accuracy | | | | |
|---|---|---|---|---|---|
| Metric | Mean Train | Mean Test | Delta: Train-Test | SD Train | SD Test |
| 2D CDF RMSE | 78.7% | 75.7% | 3.0% | 0.03% | 2.8% |
| MSE | 77.7% | 74.4% | 3.3% | 0.02% | 2.3% |
| Cosine Distance | 77.4% | 76.2% | 1.3% | 0.02% | 3.5% |

Table 6. Results of 5-times repeated stratified 10-fold cross-validation (102 pairs)

The expected drop in accuracy across the test subsets provides reasonable evidence against overfitting, especially given the limited number of observations.

## 12. Does the Size-adaptive Tuning Too Optimistic?

The complexity of the identification methods has been evaluated addressing robustness, overfitting, and generalization against achieved accuracies applying two methods of hyperparameter tuning: simple and size-adaptive.

The simple tuning of the top four AAG methods optimizes only two hyperparameters that maximizes the weighted accuracy of the identification results. However, switching into the size-adaptive tuning requires optimization of five hyperparameters as been shown in the example above: a split point of the sample size, and two sets of grid sizes *k* and smoothing parameters *bw_par* (see Table 3). It means that the size-adaptive tuning has lower degrees of freedom adjusting five hyperparameters.

By analogy to the degrees of freedom adjusted $R^2$ statistic, degrees of freedom adjusted weighted accuracy has been considered (Theil, 1961):

$$Acc_{adj}^{W} = 1 - (1 - Acc^{W})\frac{N_{eff}}{N_{eff} - n}$$

where $Acc^{W}$ is weighted accuracy, $N_{eff}$ is Kish's effective sample size, and $n$ is number of hyperparameters (i.e., $n$ equals 2 in case of the simple tuning or 5 for size-adaptive tuning).

Kish's effective sample size addresses weights contribution (Kish, 1992):

$$N_{eff} = \frac{(\sum_{i=1}^{102} W_i)^2}{\sum_{i=1}^{102} W_i^2}$$

In our case, it significantly reduces freedom of 102 pairs down to 56.7.

Across three top metrics, the adjusted accuracies of the size-adaptive tuning outperform simply tuned results (Table 7).

| | Weighted Accuracy | | | Adjusted Accuracy | | | AIC | | | BIC | | |
|---|---|---|---|---|---|---|---|---|---|---|---|---|
| Tuning | Simple | | Size-Adaptive | Simple | | Size-Adaptive | Simple | | Size-Adaptive | Simple | | Size-Adaptive |
| # of Hyperparameters | 2 | | 5 | 2 | | 5 | 2 | | 5 | 2 | | 5 |
| 2D CDF RMSE | 78.6% | → | 80.1% | 77.8% | → | 78.2% | -83.5 | ← | -81.7 | -79.4 | ← | -71.5 |
| MSE | 77.7% | → | 80.4% | 76.9% | → | 78.6% | -81.0 | → | -82.6 | -77.0 | ← | -72.4 |
| Cosine Distance | 77.4% | → | 79.8% | 76.6% | → | 77.8% | -80.5 | → | -80.6 | -76.4 | ← | -70.4 |

Table 7. Simple versus size-adaptive tuning comparisons

To further address complexity of the tuning against obtained accuracy and generalization, Akaike Information Criterion (AIC) and Bayesian Information Criterion (BIC) have been estimated (Burnham and Anderson, 2002). These two criteria reveal controversial but informative results across three metrics. According to AIC the size-adaptive tuning is preferable for MSE and cosine distance while stricter BIC controverts that conclusion. This represents a borderline case; more real-world cause-effect data would facilitate a more decisive conclusion.

## 13. Screening of Indecisive Cases

The direction of causation may be addressed with the preceding clarification: How decisive is the method to identify causal directionality by observing some bivariate data statistics? Knowing the answer to that question may help screen out indecisive cases prior to running the identification of causal directionality.

For example, the simply tuned AAG method with applied cosine distance or Pearson correlation metrics incorrectly identifies 24 cases out of 102 pairs (Appendix, Figure A3). This information can be used to set an additional binary label (correct=0 and incorrect=1) to the data pairs and run a classification model providing descriptive statistics as input features. The following symmetrical 2D statistics of $X$ and $Y$ have been selected as input features: mutual information (*sklearn.metrics.mutual_info_score*), Kendall, Pearson, and Spearman correlations and $R^2$, cosine distance, Jaccard index, and diagonal symmetry score. The latter has been calculated using *stats.gaussian_kde*.

A simple shallow decision tree *DecisionTreeClassifier* from *sklearn.tree* has been fitted providing interpretable classification (Figure 7).

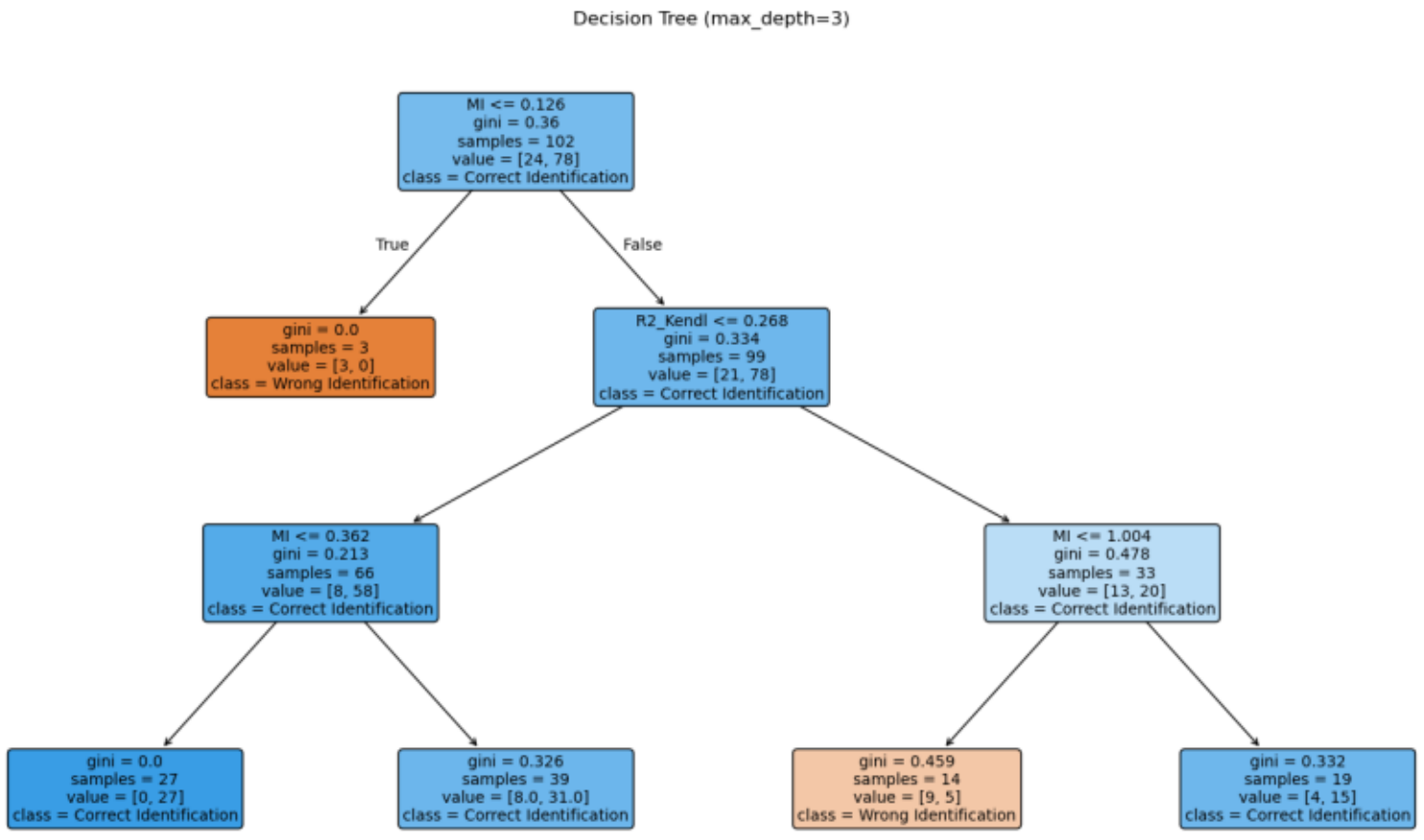

Figure 7. Example of a decision tree (ROC AUC=0.81, 102 pairs)

It turns out that two statistics: mutual information and Kendall $R^2$ are important features to screen out pairs that will most likely be incorrectly classified by the AAG method (Figure 8).

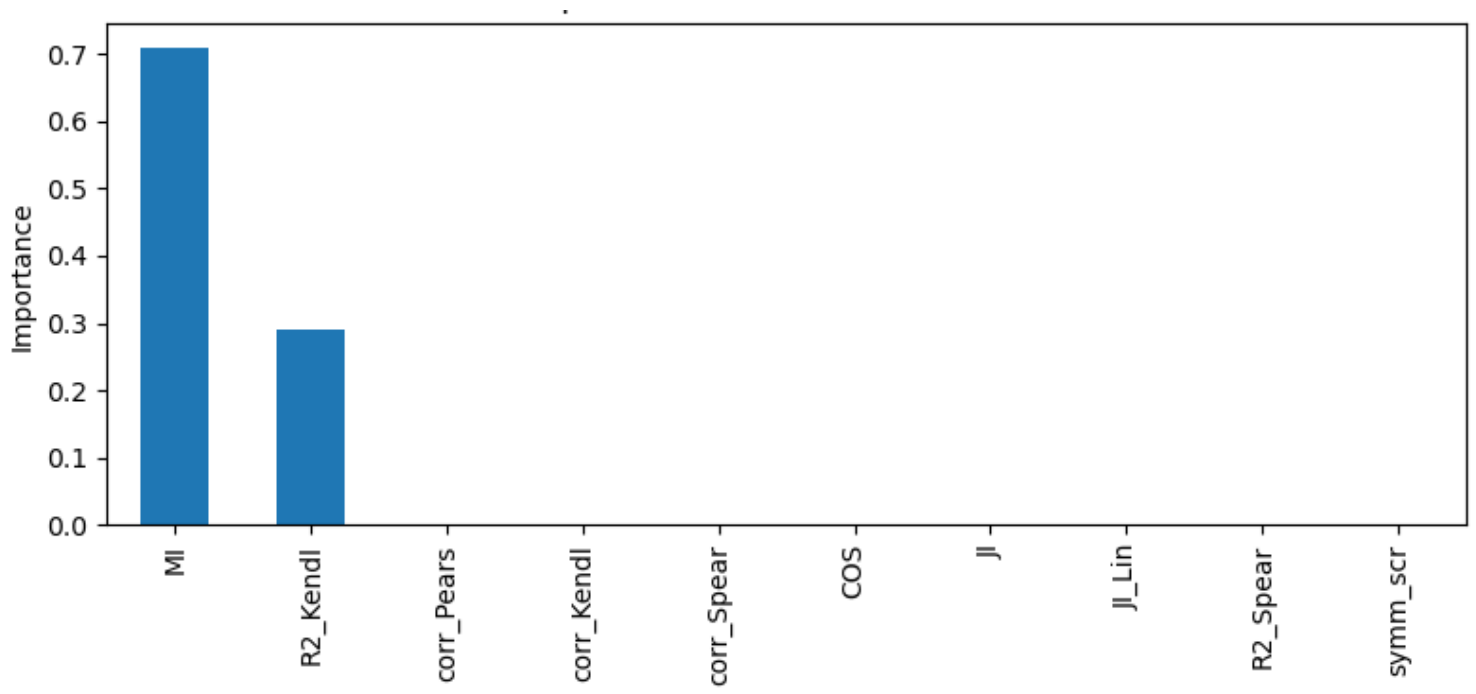

Figure 8. Input feature importance of the fitted decision tree model

According to the fitted decision tree, three pairs (IDs: 17, 87, 107) having mutual information less than 0.125 have been isolated out of 24 as indecisive. Identification of causal directionality by AAG for the remaining 99 pairs achieved a significantly higher weighted accuracy of about +6% after either simple or size-adaptive tuning (Table 4).

Notably, other top performing AAG methods applying 2D CDF RMSE or MSE metrics also misclassified these three pairs. Furthermore, when applying the MSE metric, the misclassified cases intersect strongly with the cosine distance results (23 out of 25, or 92%). In contrast, the 2D CDF RMSE metric shows a lower overlap, matching 76% of the pairs.

## 14. Future Studies

The presented study has been limited to 102 pairs of the Tübingen real-world examples all with known causations to explore AAG and Monotonicity Index approaches. It shows quite promising results for AAG method. Having more data, especially of different types, statistical characteristics, and with different dependencies, will provide an opportunity to enhance the approach.

Future research may address preceding classification between cases with and without cause-effect relationships. Given a pair of numeric data, the chain of questions is: (1) Does a bivariate cause-effect relationship exist? and if the answer is “Yes”, then (2) What is the direction of causation? The first classification is an extremely challenging multivariable problem that is out of scope of this study.

Switching from binary distinguishing (3) to more informative values of metrics **M**, it may provide additional lift. Furthermore, enhancing metrics **M** by symmetrical statistics, as described in Section 13, and then fitting machine learning classification models, it may boost accuracy even further.

In addition, it is reasonable to explore ensemble methods, such as AAG with different metrics or incorporating AAG with other approaches, such as ANM or GRCI.

Future studies may include different geometries of anticipated distributions in addition to the described dual response with normally distributed fitting (1). While uniform distributions and a triple response approach have also been considered, both failed to yield satisfactory results. The latter has been performed by utilizing *stats.johnsonsu.pdf* distribution using three moments: mean, standard deviation, and skewness. Nevertheless, it may be a subject of a deeper study, including, for example, unimodal asymmetrical distributions or higher moments, such as kurtosis. Potential challenges of that study are issues associated with more synthetic distributions compared to the real-world or compromised robustness of the results incorporating more complex methods.

## 15. Conclusions

The paper focuses on the identification of causal directionality by considering conditional distributions and features two methods: (1) Anticipated Asymmetric Geometries (AAG) and (2) Monotonicity Index.

The presented study has been limited to the Tübingen real-world 102 pairs of examples all with known causations to explore AAG and Monotonicity Index methods. It shows quite promising results for AAG with 2D CDF RMSE, followed by MSE, and cosine distance metrics. Having more data, especially of different types, statistical characteristics, and with different dependencies will provide an opportunity to enhance the approach.

Both methods assume stochastic properties of bivariate data and exploit the anticipated unimodality of conditional distributions of the effect.

In the AAG method, observation of deviations of the empirical conditional distributions from anticipated unimodal geometries of the effect, breaks the symmetry and supports discovery of causal directionality. The deviation metrics have been estimated by applying seventeen alternative approaches: 2D CDF RMSE, 2D K-S maximum distance, 2D Wasserstein, cosine distance, entropy, Hellinger distance, Jaccard index, Jeffreys divergence, J-S divergence, K-L divergence, maximum density, MaxAE, MAE, MSE, mutual information, normalized mutual information, and Pearson correlation.

Regularizations have been applied in the Monotonicity Index method to reduce impacts of instabilities of distributions gradients, especially on tails.

It was observed that the top performing method is AAG with 2D CDF RMSE metric followed by MSE, and cosine distance. These three approaches are robust, simple to calculate, and free of logarithmic functions that trigger domain errors when probability densities equal zero.

The described AAG and Monotonicity Index methods include a limited number of hyperparameters that impact the accuracy of the causal directionality identifications. For a given set of hyperparameters, the outcome of either AAG or Monotonicity Index methods are completely repeatable. Acknowledging results sensitivity to hyperparameters, tuning of hyperparameters has been done by implying full factorial Design of Experiment. Two tuning types were analyzed: simple and size-adaptive.

Size-adaptive tuning based on a single split of pairs along sample sizes with split point of 350 observations noticeably increases accuracy of AAG method by about 2.2%.

Benchmarking on the Tübingen cause-effect 95 pairs demonstrates that the AAG method with 2D CDF RMSE metric achieves 80.8% weighted accuracy after simple tuning and 82.4% after size-adaptive tuning, outperforming the ANM's 63 ± 10 %. A 99-pair comparison reveals that the AAG yields 82.7% (simple tuning) and 84.4% (size-adaptive tuning) versus 81.6% provided by GRCI or 82.0% achieved by CAREFL-H.

The AAG method is a straightforward approach: once the empirical 2D probability density function is fitted, the method produces a deterministic outcome without requiring additional model optimization or convergence of an iterative process.

The complexity of the identification methods has been assessed alongside their accuracies, robustness, potential for overfitting, and generalization performance. Across three top metrics, the degrees of freedom adjusted accuracies of the size-adaptive tuning outperform simply tuned results. According to Akaike Information Criterion (AIC), the size-adaptive tuning is preferable for MSE and cosine distance while stricter Bayesian Information Criterion (BIC) controverts that conclusion. This represents a borderline case; more real-world cause-effect data would facilitate a more decisive conclusion.

In addition, 5-times repeated stratified 10-fold cross-validation with hyperparameter tuning has been performed to address out-of-sample generalization that may expose overfitting. The expected drop in accuracy across the test subsets provides reasonable evidence against overfitting, especially given the limited number of observations.

The direction of causation may be addressed with the preceding clarification: How decisive is the method to identify causal directionality by observing some statistics of the data? Knowing the answer to that question may help screen out indecisive cases prior to running the identification of causal directionality. It turns out that the mutual information below 0.125 is the most important feature to screen out pairs that will most likely be incorrectly classified by the AAG method.

# Appendix

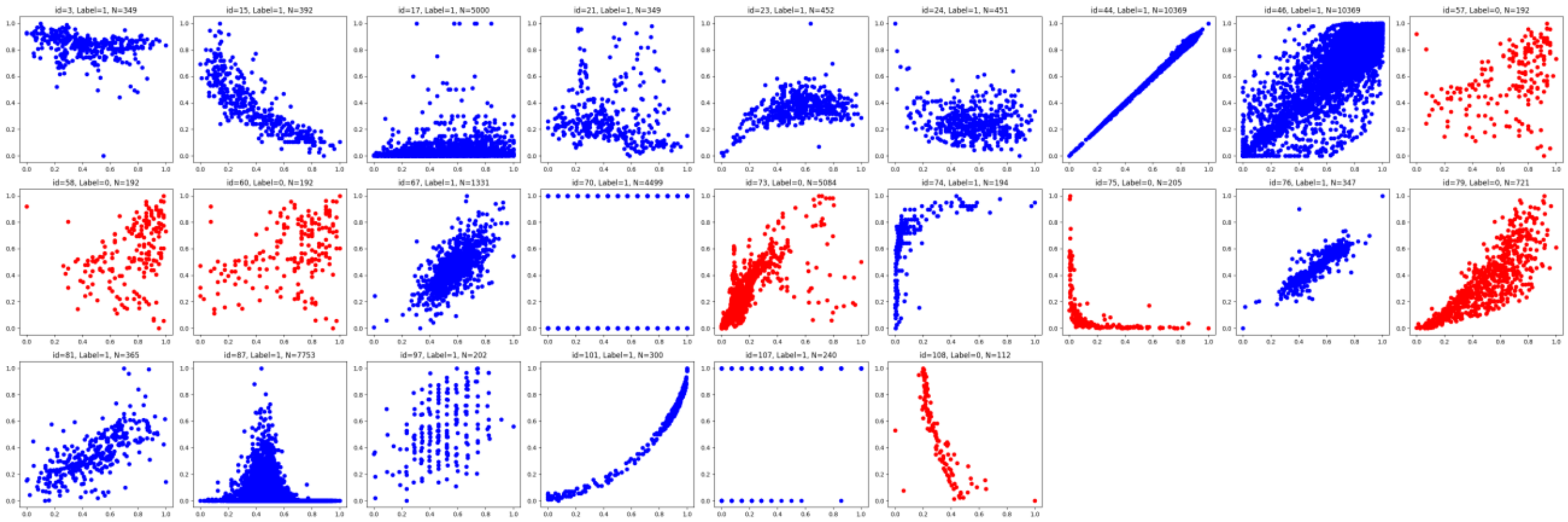

Figure A1. Scatterplots of misclassified 24 pairs by simply tuned AAG applying cosine distance metric